\documentclass[a4paper]{article}

\usepackage{INTERSPEECH2015}

\usepackage{graphicx}
\usepackage{amssymb,amsmath,bm}
\usepackage{textcomp}

\usepackage{multirow}
\usepackage{hhline}
\usepackage{multicol, blindtext}
\usepackage{array}
\newcolumntype{L}[1]{>{\raggedright\let\newline\\\arraybackslash\hspace{0pt}}m{#1}}
\newcolumntype{C}[1]{>{\centering\let\newline\\\arraybackslash\hspace{0pt}}m{#1}}
\newcolumntype{R}[1]{>{\raggedleft\let\newline\\\arraybackslash\hspace{0pt}}m{#1}}

\usepackage{placeins}
\usepackage{float}

\usepackage{makecell}
\usepackage{array}
\usepackage{caption}
\usepackage{lipsum}

\ninept

\title{ \fontsize{13}{7}\selectfont A Multi-layered Acoustic Tokenizing Deep Neural Network (MAT-DNN) for Unsupervised Discovery of Linguistic Units and Generation of High Quality Features}

\makeatletter
\def\name#1{\gdef\@name{#1\\}}

\makeatother \name{{ \em
Cheng-Tao Chung$^{*1}$, Cheng-Yu Tsai$^{\#2}$, Hsiang-Hung Lu$^{\#3}$,}\\{ \em  Yuan-ming Liou$^{\#4}$,  Yen-Chen Wu$^{\#5}$, Yen-Ju Lu$^{\#6}$, Hung-yi Lee$^{\#7}$ and Lin-shan Lee$^{\#8}$}}

\address{
Graduate Institute of Electical Engineering, National Taiwan University$^*$ \\
  Graduate Institute of Communication Engineering, National Taiwan University$^\#$ \\
{\scriptsize \tt f01921031@ntu.edu.tw$^1$, r02942067@ntu.edu.tw$^2$, r03942039@ntu.edu.tw$^3$,  qxesqxes@gmail.com$^4$,} \\ {\scriptsize \tt  r03942044@ntu.edu.tw$^5$, r03942063@ntu.edu.tw$^6$, tlkagkb93901106@gmail.com$^7$, lslee@gate.sinica.edu.tw$^8$}}

\begin{document}
\fontsize{8.26}{8}\selectfont

\maketitle
  \begin{abstract}
    This paper summarizes the work done by the authors for the Zero Resource Speech
Challenge organized in the technical program of Interspeech 2015. The goal of
the challenge is to discover linguistic units directly from unlabeled speech
data. The Multi-layered Acoustic Tokenizer (MAT) proposed in this work
automatically discovers multiple sets of acoustic tokens from the given corpus.
Each acoustic token set is specified by a set of hyperparameters that describe
the model configuration. These sets of acoustic tokens carry different
characteristics of the given corpus and the language behind thus can be
mutually reinforced. The multiple sets of token labels are then used as the
targets of a Multi-target DNN (MDNN) trained on low-level acoustic features.
Bottleneck features extracted from the MDNN are used as feedback for the MAT
and the MDNN itself. We call this iterative system the Multi-layered Acoustic
Tokenizing Deep Neural Network (MAT-DNN) which generates
high quality features for track 1 of the challenge and acoustic tokens for
track 2 of the challenge.
  \end{abstract}

\section{Introduction}
Human infants acquire knowledge of a language by mere immersion in a language speaking community. The process is not yet completely understood, and is difficult to be reproduced by current automatic speech recognition (ASR) technologies where the dominant paradigm is supervised learning with large human-annotated data sets\cite{hinton2012deep}. The idea behind the Zero Resource Speech Challenge is to inspire the development of speech recognition under the extreme situation where a whole language has to be learned from scratch\cite{lee2012nonparametric,siu2014unsupervised}. 
The goal of this challenge is to find linguistic units directly from raw audio with no knowledge of the language, the speaker, or any other supplementary information. This challenge includes two tracks which focuses on subword units and word units respectively. In the first track of unsupervised subword modeling, the aim is to construct a framewise feature representation of speech sounds, that is robust to within-speaker and across-speaker variation. Dynamic Time Warping (DTW) is performed on sequences of these features for predefined phone pair intervals to extract the warping distance. The performance of the feature is evaluated using the ABX discriminability \cite{schatz2013evaluating} on within and across-speaker phone pairs. The second track focuses on discovery of word units and the aim is to extract timing information of such word units in the hypothesized vocabularies derived from the speech corpus. The intervals in which each word unit appears in the corpus is then evaluated on parsing, clustering and matching quality \cite{ludusan2014bridging}. This paper serves as the documentation for the work by a team organized in National Taiwan University submitted to the challenge within the Interspeech 2015 technical program.

In this work, we propose a completely unsupervised framework of Multi-layered Acoustic Tokenizing Deep Neural Network (MAT-DNN) for the task. A Multi-layered Acoustic Tokenizer (MAT) is used to generate multiple sets of acoustic tokens. Each acoustic token set is specified by a pair of hyperparameters representing model granularities of the tokens. As a naming convention, we call an acoustic token set obtained from a hyperparameter pair a layer. Each layer carries complementary knowledge about the corpus and the language behind\cite{chung2014unsupervised}. Since it is well known that speech signals have multi-level structures including at least phonemes and words which are helpful in analysing or decoding speech \cite{pan2010performance}, these sets of acoustic tokens can be further mutually reinforced\cite{chung2015enhancing}. The multi-layered token labels generated by the MAT are then used as the training targets of a Multi-target Deep Neural Network\cite{vu2014investigating} (MDNN) to learn the framewise bottleneck features\cite{vesely2012language} (BNFs). The BNFs are then used as feedback to both the MAT and the MDNN in the next iteration. The BNFs from  the MDNN are evaluated in Track 1, while the time intervals for acoustic tokens obtained in the MAT are evaluated in Track 2.



\section{Proposed Approach}

\subsection{Overview of the proposed framework}
\begin{figure*}[tbh]
\centerline{\includegraphics[width=1.0\textwidth]{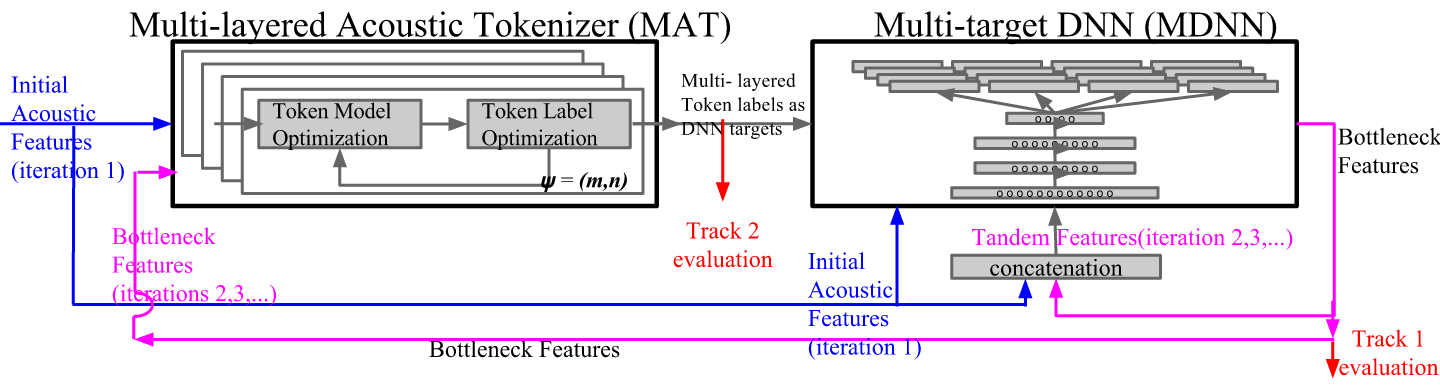}}
\caption{The proposed framework of Multi-layered Acoustic Tokenizing Deep Neural Network (MAT-DNN)}\label{fig:1}
\end{figure*}

The framework of the approach is shown in Fig\ref{fig:1}. In the left part, the Multi-layered Acoustic Tokenizer (MAT) produces many sets of acoustic tokens using unsupervised HMMs, each describing different aspects of the given corpus. These tokens are specified by two hyperparameters describing HMM configurations. A set of acoustic tokens is obtained for each configuration by iteratively optimizing the token models and the token labels on the given acoustic corpus. Multiple pairs of hyperparameters were selected producing multi-layered token labels for the given corpus to be used as the training targets of the Multi-target Deep Neural Network (MDNN) on the right part of Fig.\ref{fig:1}. The MDNN on the right learns its parameters based on the multi-layered token labels for the given corpus as its targets from the MAT on the left, so the knowledge carried by different token sets on different layers are fused. Bottleneck features are then extracted from this MDNN. In the first iteration, some initial acoustic features are used for both the MAT and the MDNN. This gives the first set of bottleneck features. These bottleneck features are then used as feedback to both the MAT (to replace the initial acoustic features) and the MDNN (to be concatenated with the initial acoustic features to produce tandem features) in the second iteration. Such feedback can be continued iteratively. The complete framework is referred to  as Multi-layered Acoustic Tokenizing Deep Neural Network (MAT-DNN) in this paper. The output of the MDNN (bottleneck features) is evaluated in Track 1 of the Challenge, while the time intervals for the acoustic token labels at the output of the MAT are evaluated in Track 2 of the Challenge. 

\subsection{Multi-layered Acoustic Tokenizer}
The goal in this step is to obtain  multiple sets of acoustic tokens, each defined by some hyperparameters, which capture complementary aspects of the corpus. There is no knowledge regarding the corpus at all, so the process here is completely unsupervised.

\subsubsection{Unsupervised Token Discovery for Each layer of MAT}
\label{sec:2-2-1}
Using unsupervised HMMs, it is straight forward to discover acoustic tokens from the corpus for a chosen hyperparameter pair $\psi$ that determines the HMM configuration (number of states per model and number of distinct models) \cite{jansen2011towards,gish2009unsupervised,siu2010improved,chung2013unsupervised,creutz2007unsupervised}.
This can be achieved by first finding an initial label set $\omega_0$ based on a set of assumed tokens for all features in the corpus $X$ as in (\ref{eq:1}) \cite{chung2013unsupervised}.
Then in each iteration $t$ the HMM parameters $\theta^\psi_{t}$ can be trained with the label set $\omega_{t-1}$ obtained in the previous iteration as in (\ref{eq:2}), and the new label set $\omega_{t}$ can be obtained by token decoding with the obtained parameters $\theta^\psi_{t}$ as in (\ref{eq:3}). 
\begin{eqnarray}
\omega_{0}&=& \mbox{initialization}(X),\phantom{\arg \max_{\substack{\theta^\psi}}}                                           \label{eq:1} \\ 
\theta^\psi_{t} &=& \arg \max_{\substack{\theta^\psi}} P(X|\theta^\psi,\omega_{t-1}),             \label{eq:2} \\
\omega_{t} &=& \arg \max_{\substack{\omega}} P(X|\theta^\psi_{t} ,\omega).                        \label{eq:3}
\end{eqnarray}
The training process can be repeated with enough number of iterations until a converged set of token HMMs is obtained. The processes (\ref{eq:2}),(\ref{eq:3}) are referred to as token model optimization and token label optimization in the left part of Fig.\ref{fig:1}.

\subsubsection{Granularity Space of Multi-layered Acoustic Token Sets}\label{sec:2-2-2}

\begin{figure}[h]
\centerline{\includegraphics[width=0.4\textwidth]{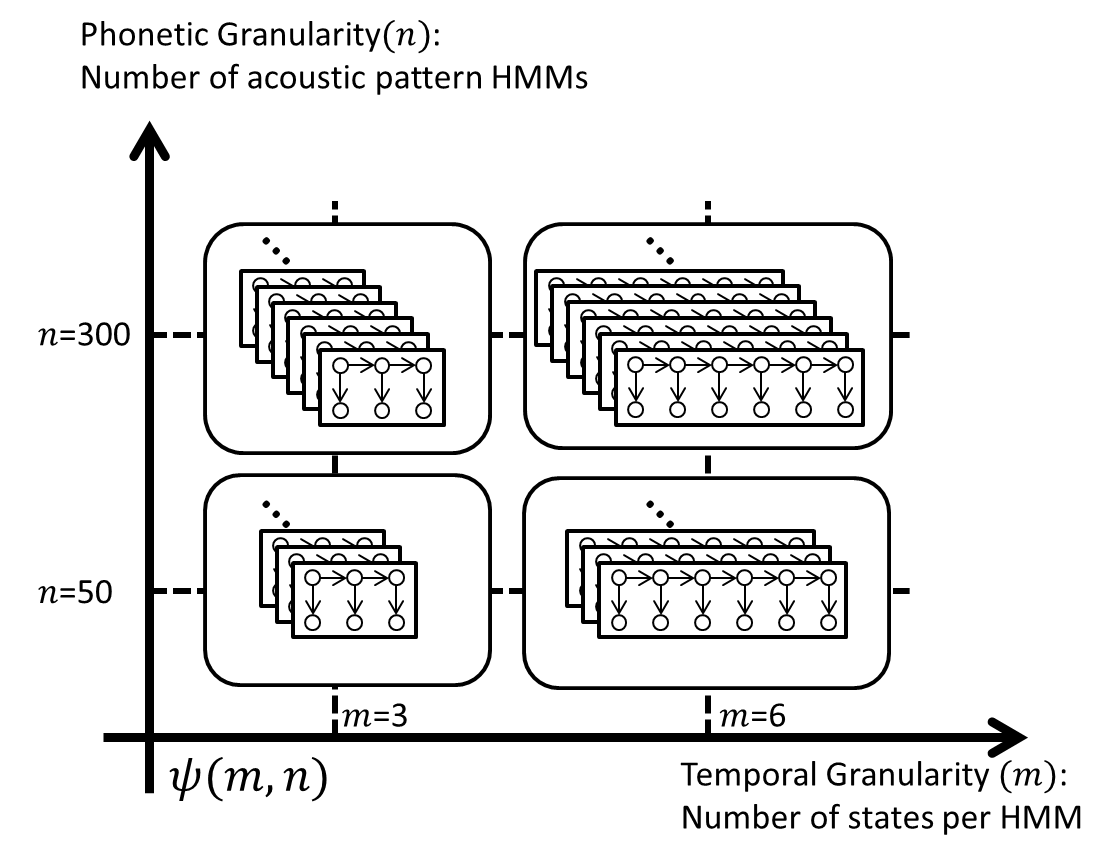}}
\caption{Model granularity space for HMM configurations}\label{fig:2dcube}
\end{figure}

The process explained above can be performed with different HMM configurations, each characterized by two hyperparameters: the number of states $m$ in each acoustic token HMM, and the total number of distinct acoustic tokens $n$ during initialization,  $\psi=(m,n)$. The transcription of a signal decoded with these tokens can be considered as a temporal segmentation of the signal, so the HMM length (or number of states in each HMM) $m$ represents the temporal granularity. The set of all distinct acoustic tokens can be considered as a segmentation of the phonetic space, so the total number $n$ of distinct acoustic tokens represents the phonetic granularity. 
This gives a two-dimensional representation of the acoustic token configurations in terms of temporal and phonetic granularities as in Fig.\ref{fig:2dcube}. Any point in this two-dimensional space in Fig.\ref{fig:2dcube} corresponds to an acoustic token configuration. 
Acoustic tokens in different layers have different model granularities that extract complementary characteristics of the corpus and the language behind, so they jointly capture knowledge about the corpus.
Although the selection of the hyperparameters can be arbitrary in the above two-dimensional space, here we can select $M$ temporal granularities ($m$=$m_1$,$m_2$,...$m_M$) and $N$ phonetic granularities ($n$=$n_1$,$n_2$,...$n_N$), forming a two-dimensional array of $M \times N$ hyperparameter pairs in the granularity space.

\subsection{Mutual Reinforcement of Multi-layered Tokens}
Because all the layers obtained in the MAT above are learned in an unsupervised fashion, they are not precise. But we have many layers, each corresponding to a different pair of hyperparameters $\psi=(m,n)$, so they can be mutually reinforced. This is explained here and shown in Fig.\ref{fig:3}, including token boundary fusion and LDA-based token label re-initialization as in Fig.\ref{fig:3}(a).

\subsubsection{Token Boundary Fusion}
Fig.\ref{fig:3}(b) shows the token boundary when a part of an utterance is segmented into acoustic tokens on different layers with different hyperparameter pairs $\psi=(m,n)$. We define a boundary function $b_{m,n}(j)$ on each layer with $\psi=(m,n)$ for the possible boundary between every pair of two adjacent frames within the utterance, where $j$ is the time index of such possible boundaries. On each layer $b_{m,n}(j)$=1 if boundary $j$ is a token boundary and 0 otherwise. All these boundary functions $b_{m,n}(j)$ for all different layers are then weighted and averaged to give a joint boundary function $B(j)$. The weights consider the fact that smaller $m$ or shorter HMMs generate more boundaries. The peaks of $B(j)$ are then selected based on the second derivatives and some filtering and thresholding process. This gives the new segmentation of the utterance as shown at the bottom of Fig.\ref{fig:3}(b).

\subsubsection{LDA-based Token Label Re-initialization}
\begin{figure}[tbh]
\centerline{\includegraphics[width=0.48\textwidth]{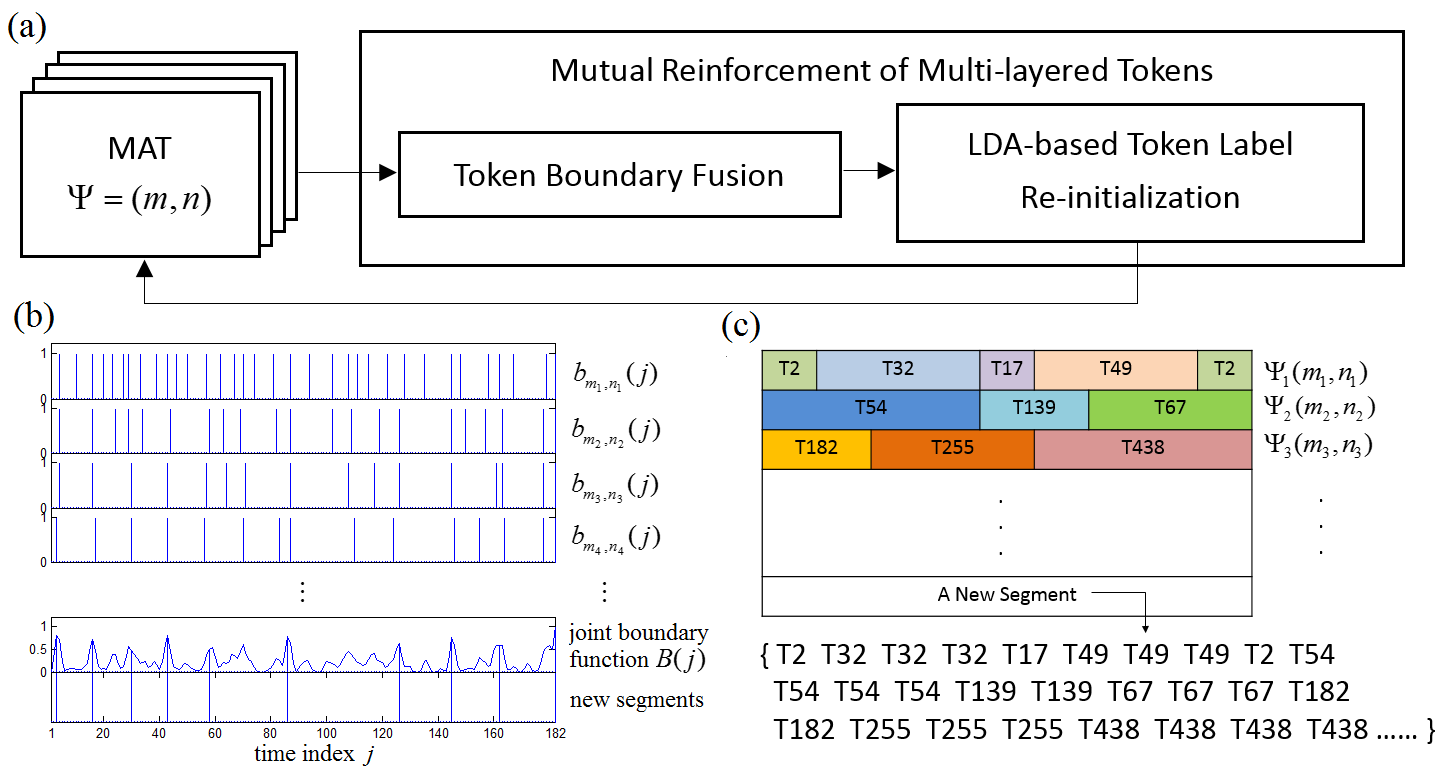}}
\caption{Mutual reinforcement of multi-layered tokens: (a) block diagram, (b) token boundary fusion, and (c) a new segment considered as a document (bag-of-words) and a token as a word in LDA based token label re-initialization.
}\label{fig:3}
\end{figure}

As shown in Fig.\ref{fig:3}(c), each new segment obtained above usually consists of a sequence of acoustic tokens on each layer based on the tokens defined on that layer. We now consider all the tokens on all the different layers as different words, so we have a vocabulary of $\sum\limits_{i=1}^{MN}     n_i$ words, i.e., there are $n_i$ words on the $i$-th layer and there are a total of $MN$ layers. A new segment here is thus considered as a document (bag-of-words) composed of words (tokens) collected from all different layers. Latent Dirichlet Allocation\cite{blei2003latent} (LDA) is preformed for topic modeling, and then each document (new segment) is labeled with the most probable topic. Because in LDA a topic is characterized by a word distribution, here a token distribution across different layers may also represent a certain acoustic characteristics or a certain acoustic token. By setting the number of topics in LDA as the number of distinct tokens $n$ ($n$=$n_1$,$n_2$,...$n_N$) as in subsection \ref{sec:2-2-2}) we have a new initial label set $\omega_0$ as in (\ref{eq:1}) of subsection \ref{sec:2-2-1}, in which each new segment obtained here is a new acoustic token whose ID is the topic ID obtained by LDA. This new initial label set $\omega_0$ is then used to re-train all the acoustic tokens on all layers of MAT as in (\ref{eq:1})(\ref{eq:2})(\ref{eq:3}).

\subsection{The Multi-target DNN (MDNN)}
As shown in the right part of Fig.\ref{fig:1}, token label sequence from a layer (with a pair of hyperparameters $\psi=(m,n)$) is a valid target for supervised framewise training, although obtained in an unsupervised way. In the initial work here, we do not use the HMM states as the target, but simply take the token label as the training target. As shown in Fig.\ref{fig:1}, there are multi-layered token labels with different hyperparameter pair $\psi=(m,n)$ for each utterance, so we jointly consider all the multi-layered token labels by learning the parameters for a single DNN with a uniformly weighted cross-entropy objective at the output layer. As a result, the bottleneck feature (BNF) extracted from this DNN automatically fuse all knowledge about the corpus and the language behind learned from the different sets of acoustic tokens.


\begin {table}[t]

\begin{center}
\begin{tabular}{|c|l|p{0.033\textwidth}|p{0.033\textwidth}|p{0.033\textwidth}|p{0.033\textwidth}|}
\hline
\multicolumn{2}{|c|}{\multirow{2}{*}{Method}} & \multicolumn{2}{c|}{English}    & \multicolumn{2}{c|}{Tsonga}     \\ \cline{3-6} 
\multicolumn{2}{|c|}{}                        & across         & within         & across         & within         \\ \hline
(1)             & Baseline                    & 28.10          & 15.60          & 33.80          & 19.10          \\ \hhline{|=|=|=|=|=|=|}
(2)             & MFCC                        & 28.63          & 15.89          & 30.77          & 16.34          \\ \hline
(3)             & DBM posterior                       & 25.96          & 15.74          & 29.15          & 16.18          \\ \hline
(4)             & BNF-1st, MR-0               & 26.84          & 15.95          & 26.48          & 15.52          \\ \hline
(5)             & BNF-1st, MR-1               & 23.88          & 14.60          & 21.97          & 13.40          \\ \hline
(6)             & BNF-1st, MR-2               & 24.46          & 14.92          & 22.14          & 13.31          \\ \hline
(7)             & BNF-2nd, MR-0               & 26.55          & 16.27          & 26.23          & 15.05          \\ \hline
(8)             & BNF-2nd, MR-1               & 24.53          & 15.13          & 23.30          & 13.88          \\ \hline
(9)             & BNF-1st, MR-1*            & \textbf{21.92} & \textbf{13.95} & \textbf{21.42} & \textbf{12.84} \\ \hline
(10)            & BNF-2nd, MR-1*            & 24.13          & 15.24          & 23.05          & 14.03          \\ \hhline{|=|=|=|=|=|=|}
(11)            & Topline                     & 16.00          & 12.10          & 04.50          & 03.50          \\ \hline
\end{tabular}
\end{center}

\caption {Results for Track 1 of the challenge, the best figure for each metric is shown in bold. \label{tab:1}}
\end{table}

\subsection{The Iterative Learning Framework for MAT-DNN}
Once the BNFs are extracted from the MDNN in iteration 1, they can be taken as the input of the MAT on the left of Fig.\ref{fig:1}(c) replacing the initial acoustic features. The MAT then generates updated sets of multi-layered token labels and these updated sets of multi-layered token labels can be used as the updated training objective of the MDNN. The input features of the MDNN can also be updated by concatenating the initial acoustic features with the newly extracted BNFs as the tandem features. This process can be repeated for several iterations until satisfactory results are obtained.  The tandem feature used as the input of the MDNN can be further augmented by concatenating unsupervised features obtained in other systems such as the Deep Boltzmann Machine\cite{salakhutdinov2009deep} (DBM) posteriorgrams, Long-Short Term Memory Recurrent Neural Network\cite{hochreiter1997long} (LSTM-RNN) autoencoder bottleneck features, and i-vectors\cite{kanagasundaram2011vector} trained on MFCC.
Although different from the conventional recurrent neural network (RNN) in which the recurrent structure is included in back propagation training, the concatenation of the bottleneck features with other features in the next iteration in MDNN is a kind of recurrent structure.

\label{sec:2-5}

\begin{table*}[h!t]
\centering
\tabcolsep=0.11cm
\begin{tabular}{!{\vrule width 1pt}c|c|l!{\vrule width 1pt}cc!{\vrule width 1pt}ccc!{\vrule width 1pt}ccc!{\vrule width 1pt}ccc!{\vrule width 1pt}ccc!{\vrule width 1pt}ccc!{\vrule width 1pt}}
\Xhline{3\arrayrulewidth}
\multicolumn{3}{!{\vrule width 1pt}c!{\vrule width 1pt}}{\multirow{2}{*}{(\%)}}                                                               & \multirow{2}{*}{NED} & \multirow{2}{*}{Cov.} & \multicolumn{3}{c!{\vrule width 1pt}}{Matching}      & \multicolumn{3}{c!{\vrule width 1pt}}{Grouping} & \multicolumn{3}{c!{\vrule width 1pt}}{Type}          & \multicolumn{3}{c!{\vrule width 1pt}}{Token}                   & \multicolumn{3}{c!{\vrule width 1pt}}{Boundary}        \\ \cline{6-20} 
\multicolumn{3}{!{\vrule width 1pt}c!{\vrule width 1pt}}{}                                                                                    &                      &                       & P    & R            & F            & P        & R        & F       & P   & R             & F            & P            & R             & F             & P    & R             & F             \\ \Xhline{3\arrayrulewidth}
\multirow{2}{*}{Eng.} & \multicolumn{2}{c!{\vrule width 1pt}}{JHU}                                                          & 21.9                 & 16.3                  & 39.4 & 1.6          & 3.1          & 21.4     & 84.6     & 33.3    & 6.2 & 1.9           & 2.9          & 5.5          & 0.4           & 0.8           & 44.1 & 4.7           & 8.6           \\ \cline{2-20} 
                      & (A) & \begin{tabular}[c]{@{}l@{}}(4) BNF-1st, MR-0\\ $\psi=(7,50)$\end{tabular}   & 87.5                 & 100                   & 1.4  & 0.5          & 0.8          & 3.6      & 18.7     & 6       & 4.2 & \textbf{11.9} & \textbf{6.2} & \textbf{8.3} & \textbf{15.7} & \textbf{10.9} & 35.2 & \textbf{84.6} & \textbf{49.8} \\ \Xhline{3\arrayrulewidth}
\multirow{3}{*}{Tso.} & \multicolumn{2}{c!{\vrule width 1pt}}{JHU}                                                          & 12                   & 16.2                  & 69.1 & 0.3          & 0.5          & 52.1     & 77.4     & 62.2    & 3.2 & 1.4           & 2            & 2.6          & 0.5           & 0.8           & 22.3 & 5.6           & 8.9           \\ \cline{2-20} 
                      & (B) & \begin{tabular}[c]{@{}l@{}}(8) BNF-2nd, MR-1\\ $\psi=(9,50)$\end{tabular}   & 69.1                 & 95                    & 5.9  & \textbf{0.5} & \textbf{0.9} & 10.7     & 26.8     & 15.3    & 1.5 & 3.9           & 2.2          & 2.3          & 6.6           & 3.4           & 17.1 & 59.1          & 26.6          \\ \cline{2-20} 
                      & (C) & \begin{tabular}[c]{@{}l@{}}(5) BNF-1st, MR-1\\ $\psi=(13,300)$\end{tabular} & 60.2                 & 96.1                  & 9.7  & 0.4          & 0.8          & 13.5     & 12.7     & 13.1    & 1.8 & \textbf{4.7}  & \textbf{2.5} & \textbf{3.9} & \textbf{9.1}  & \textbf{5.4}  & 21.2 & \textbf{62.1} & \textbf{31.6} \\ \Xhline{3\arrayrulewidth}
\end{tabular}
\caption{Comparison of three typical example token sets selected out of all shown in Fig.\ref{fig:t1} with the JHU baseline. Those better than JHU baseline are in bold.}
\label{tab:2}
\end{table*}

\section{Experimental Setup}

The general framework of the MAT-DNN presented above allows several flexible configurations. However, in this work we train the MAT-DNN in the following manner. We set $m$=3, 5, 7, 9 states per token HMM and $n$=50, 100, 300, 500 distinct tokens in the MAT, which gives a total of 16 layers.

In the first iteration, we use the 39 dimension Mel-frequency Cepstral Coefficients (MFCC) with energy, delta and double delta as the initial acoustic features for the input to both the MAT and the MDNN.  We tandem the MFCC with a window of 4 frames before and after (39x9 dimensions), and an i-vector (400 dimensions) trained on the MFCC of each evaluation interval for the input of the MDNN. The topology of the DNN is set to be 751(input)-256(hidden)-256(hidden)-39(bottleneck)-(target) with 3 hidden layers. 
Even without the feedback and tandem features, the MAT-DNN is a powerful self-contained unsupervised feature extractor. We compared the BNF extracted in the first iteration with the Deep Boltzmann Machine posteriorgrams mentioned in section \ref{sec:2-5} that use the same MFCC as input. To make the comparison fair, we keep the dimensionality of these features to be 39. For the Deep Boltzmann Machine, we used the 39-dimension MFCC with a window of 5 frames before and after as the input. The configuration we used for the DBM is 429(visible)-256(hidden)-256(hidden)-39(hidden). We originally extracted another set of LSTM-RNN autoencoder bottleneck features as another baseline but the performance was slightly worse than the MFCC thus we omit it in any discussion here.

In the second iteration, we tandem the original MFCC, the BNF extracted from the first iteration, the DBM posteriorgrams, and the i-vector forming a (39x9+39x9+39x9+400=1453) dimension input to the MDNN. We used the updated transcriptions as the target and extracted the BNF as the features. The MAT is trained using the zrst\cite{chung2014zero}, a python wrapper for the HTK toolkit\cite{young1997htk}, srilm\cite{stolcke2002srilm} that we developed for training unsupervised HMMs with varying model granularity. The LDA tool we used in the Mutual Reinforcement is done with MALLET\cite{mccallum2002mallet}. The MFCC were extracted using the HTK toolkit\cite{young1997htk}. The i-vectors were extracted using Kaldi\cite{Povey_ASRU2011}. The DBM posteriorgram is extracted using libdnn\cite{chou2014libdnn}. The MDNN was trained using  Caffe\cite{jia2014caffe}.

\subsection{Track 1}

The two official corpora are the Buckeye corpus \cite{pitt2007buckeye} and NCHLT Xitsonga Speech corpus \cite{de2014smartphone} in English and Tsonga respectively. They are used in the evaluation based on the ABX discriminability test \cite{schatz2013evaluating} including across and within speaker tests. The final results is in error percentage, which means the lower the better. Our results of track 1 is presented in Table \ref{tab:1}.

\begin{figure*}[th!]
\centerline{\includegraphics[width=\textwidth]{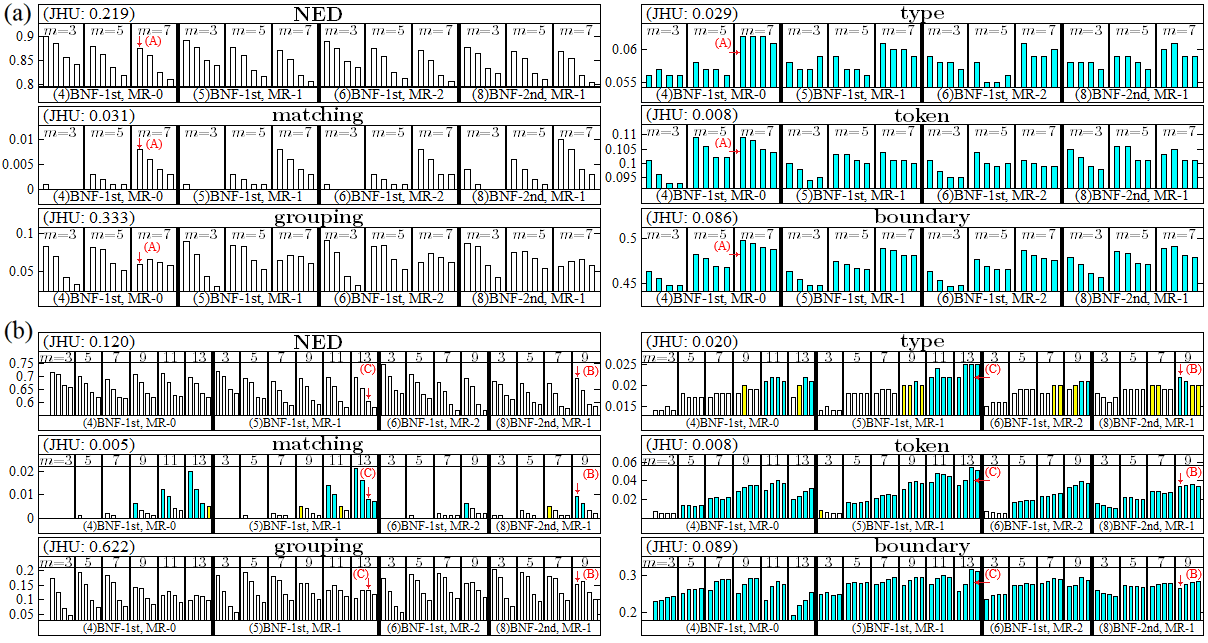}}
\caption{Results for Track 2 for (a) English and (b) Tsonga. Each subgraph is an evaluation measure for four cases of token sets used to train the bottleneck features listed in four rows of Table \ref{tab:1} as shown at the bottom. The four bars in each group for a value of $m$ are for $n$=50, 100, 300, 500 from left to right (not shown in the figure) and $\psi=(m,n)$ are parameters for the token sets. Blue, yellow and white bars correspond to better, equal to or worse as compared to the JHU baseline at the upper left corner of each subgraph. The coverage is not shown because it is almost 100\% in all cases.}
\label{fig:t1}
\end{figure*}

Rows (1) and (11) are the official baseline MFCC features and official topline supervised phone posteriorgrams provided by the challenge organizers respectively. Row (2) is our baseline of the MFCC features, the initial acoustic features used to train all systems in this work. Row (3) is for the DBM posteriorgrams extracted from the MFCC of row (2), serving as a strong unsupervised baseline. The results in rows (4), (5) and (6) are the performance of the bottleneck features extracted in the first iteration of the MAT-DNN without applying mutual reinforcement (MR) (4), applying MR once (5), and twice (6) respectively. Row (9) is similar to row (5), except we use a wider bottleneck layer with 256 dimensions instead of 39.
Rows (7) and (8) are the performance of the bottleneck features extracted in the second iteration of the MAT-DNN without applying MR (7) and applying MR once (8). The MAT of the MAT-DNN in (7) and (8) is trained using the BNF of row(5). Row (10) is similar to row (8), except only the MFCC and i-vectors are tandemed as input without other features.


All the features from row (2) to (10) except for (9) are confined to 39 dimensions. This allows fast and fair comparison of different algorithms. We observe that as a stand-alone feature extractor without any iterations, the MAT-DNN in row (5) outperforms the DBM baseline in (3). The effect of mutual reinforcement can be seen in the improvement from row (4) to row (5)(6) and row (7) to row(8). We observe that a single iteration of mutual reinforcement of the target of the MAT-DNN is enough to bring huge improvement to the system. 
The effect of iterations in the MAT-DNN can be seen by comparing rows (2), (5), (8), respectively corresponding to 0, 1, and 2 iterations. Although the performance improvement from row (2) to row (5) is notable, it dropped in the second iteration in (8). To investigate reasons of the performance drop, we widened the bottleneck feature to 256 dimensions in (9) and observed a dramatic improvement in performance. It is possible that we have not explored the full potential of the MAT-DNN as comparison between algorithms was the original goal when we designed the experiments. For a better tuned set of parameters, improvement in following iterations is to be expected on track 1. Nonetheless, the benefit of the second iteration is better observed in track 2.

\subsection{Track 2}

The evaluation tool for track 2 provided by the challenge organizers\cite{ludusan2014bridging} gives five main metrics
plus two more scores: NED and coverage.
Fig.\ref{fig:t1} shows the results for (a) English and (b) Tsonga in NED, as well as the F-measures for the five main metrics: matching, grouping, type, token, and boundary, each in a subgraph. We omit coverage here because it is almost 100\% in all cases. So there are six subfigures in Fig.\ref{fig:t1}(a) and (b). In each subfigure, the results for four cases are shown, they correspond to the four MAT targets used for the MDNN bottleneck features listed in rows (4), (5), (6) and (8) of Table \ref{tab:1}.
For each of these token sets, the three or six groups of bars correspond to different values of $m$ ($m$=3, 5, 7 or $m$=3, 5, 7, 9, 11, 13), while in each group the four bars correspond to the values of $n$ ($n$=50, 100, 300, 500 from left to right), where $\psi=(m,n)$ are the parameters for the token sets. Those bars in blue are better than the JHU baseline, while those in white are worse. Only the results jointly considering both within and across talker conditions are shown.

From Fig.\ref{fig:t1}(a) for English, it can be seen that the proposed token sets perform well in type, token and boundary scores, although much worse in matching and grouping. we see in many cases the benefits brought by MR (e.g. (6) vs (5) in type of Fig.\ref{fig:t1}(a)) and the second iteration (e.g. (8) vs (6) in boundary of Fig.\ref{fig:t1}(a)), especially for small values of $m$. In many groups for a given $m$, smaller values of $n$ seemed better, probably because $n$=50 is close to the total number of phonemes in the language. Also, a general trend is that larger values of $m$ were better, probably because HMMs with more states were better in modelling the relatively long units; this may directly lead to the higher type, token and boundary scores.

Similar observations can be made for Tsonga in Fig.\ref{fig:t1}(b), and the overall performance seemed to be even better as the proposed token sets perform well even in matching scores. The improvements brought by MR, the bottleneck features and the second iteration is better observed here, which gives the best cases for all the five main scores. This is probably due to the fact that more sets of tokens were available for  MR and MAT-DNN  on Tsonga than English. We can conclude from this observation that more token sets introduces more robustness and that leads to better token sets for the next iteration.  
When $m$ goes to 13, we see that without MR in (4) of Fig.\ref{fig:t1}(b)) almost all metrics degrade except for matching scores, but with MR almost all the scores consistently increases (except for NED) when $m$ becomes larger. This suggests that MR can also prevent degradation from happening while detecting relatively long units.

We also selected three typical example token sets (A)(B)(C) out of the many proposed here and shown in Fig.\ref{fig:t1}, and compared them with the JHU baseline\cite{jansen2011efficient} in Table \ref{tab:2} including Precision (P), Recall (R) and F-scores (F). These three example sets are also marked in Fig.\ref{fig:t1}. In Table \ref{tab:2} those better than JHU baseline are in bold. The much higher NED and coverage scores suggest that the proposed approach is a highly permissive matching algorithm. The much higher parsing scores (type, token and boundary scores), especially the Recall and F-scores, imply the proposed approach is more successful in discovering word-like units. However, the matching and grouping scores are much worse probably because the discovered tokens cover almost the whole corpus, including short pauses or silence, and therefore many tokens are actually noises. Another possible reason might be that the values of $n$ used are much smaller than the size of the real word vocabulary, making the same token label used for signal segments of varying characteristics and this degenerated the grouping qualities.

\section{Conclusion}
This paper summarizes the preliminary work done for the Zero Resource Speech Challenge in Interspeech 2015. We propose a MAT-DNN to generate multi-layer token sets and fuse the various knowledge in different token sets in the bottleneck features. We present the complete results on all evaluations we tested up to the submission deadline, with a hope that these results serve as good references for future investigations.

\FloatBarrier
  \newpage
  \eightpt
\fontsize{8.5}{11}\selectfont    
  \bibliographystyle{IEEEtran}

  \bibliography{mybib}


\end{document}